\documentclass[11pt,letterpaper]{article}
\usepackage{acl-hlt2011}
\usepackage{times}
\usepackage{latexsym}
\usepackage{pstricks,epsfig,amsmath,amssymb,multirow,color}
\usepackage{url}
\usepackage{arydshln}

\newcommand{\argmin}{\operatornamewithlimits{arg\,min}}

\definecolor{newgreen}{rgb}{0,0.5,0}

\title{
Sparse Regression for Machine Translation} 

\author{Ergun Bi\c{c}ici\\
 {Department of Electrical and Computer Engineering} \\
  Ko\c{c} University \\ 
  34450 Sariyer, Istanbul, Turkey \\
  {\tt  ebicici@ku.edu.tr}
}

\date{}

\begin{document}

\maketitle

\begin{abstract}
We use transductive regression techniques to learn mappings between source and target features of given parallel corpora and use these mappings to generate machine translation outputs. 
We show the effectiveness of $L_1$ regularized regression (\textit{lasso}) to learn the mappings between sparsely observed feature sets versus $L_2$ regularized regression.
Proper selection of training instances plays an important role to learn correct feature mappings within limited computational resources and at expected accuracy levels. 
We introduce \textit{dice} instance selection method for proper selection of training instances, which plays an important role to learn correct feature mappings for improving the source and target coverage of the training set. 
We show that $L_1$ regularized regression performs better than $L_2$ regularized regression both in regression measurements and in the translation experiments using graph decoding. We present encouraging results when translating from German to English and Spanish to English.
We also demonstrate results when the phrase table of a phrase-based decoder is replaced with the mappings we find with the regression model.
\end{abstract}

\section{Introduction}

Regression can be used to find mappings between the source and target feature sets derived from given parallel corpora. 
Transduction learning uses a subset of the training examples that are closely related to the test set without using the model induced by the full training set. 
In the context of statistical machine translation, translations are performed at the sentence level and this enables us to select a few training instances for each test instance to guide the translation process. This also gives us a computational advantage when considering the high dimensionality of the problem.

The goal in transductive regression based machine translation (RegMT) is both reducing the computational burden of the regression approach by reducing the dimensionality of the training set and the feature set and also improving the translation quality by using transduction. 
In an idealized feature mapping matrix where features are word sequences, we would like to observe few target features for each source feature 
close to permutation matrices with one nonzero item for each column. $L_1$ regularization helps us achieve solutions close to the permutation matrices by increasing sparsity.

We compare $L_1$ regularized regression with other regression techniques and show that it achieves better performance in estimating target features and in generating the translations during graph decoding.
We present encouraging results on German and Spanish to English translation tasks. 
We also replace the phrase table of a phrase-based decoder, Moses~\cite{Koehn07:moses}, with the feature mappings we find with the RegMT model.




\textbf{Related Work:} Regression techniques can be used to model the relationship between strings~\cite{CortesS2S2007}. Wang et al.~\shortcite{wang-shawetaylor-szedmak:2007:ShortPapers} applies a string-to-string mapping approach to machine translation by using ordinary least squares regression and $n$-gram string kernels to a small dataset. Later they use $L_2$ regularized least squares regression~\cite{wang-shawetaylor:2008:WMT}. Although the translation quality they achieve is not better than Moses~\cite{Koehn07:moses}, which is accepted to be the state-of-the-art, they show the feasibility of the approach. Serrano et al.~\shortcite{Serrano2009} use kernel regression to find translation mappings from source to target feature vectors and experiment with translating hotel front desk requests. 
Locally weighted regression solves separate weighted least squares problems for each instance~\cite{Hastie2009}, weighted by a kernel similarity function. 

\textbf{Outline:} Section 2 gives an overview of regression based machine translation, which is used to find the mappings between the source and target features of the training set. We present $L_1$ regularized transductive regression for alignment learning and the instance selection method used. In section 3, we compare the performance of $L_1$ versus $L_2$ regularized regression results in effectively estimating the target features. 
In section 4, we present the graph decoding results on German-English and Spanish-English translation tasks. Section 5 compares the results obtained with Moses with different learning settings. The last sections present our contributions. 


\section{Machine Translation Using Regression}
\label{RBMT}

Let X and Y correspond to the sets of tokens that can be used in the source and target strings, then, $m$ training instances are represented as $(\textbf{x}_1, \textbf{y}_1), \ldots , (\textbf{x}_m, \textbf{y}_m) \in X^* \times Y^*,$
where $(\textbf{x}_i, \textbf{y}_i)$ corresponds to a pair of source and target language token sequences for $1 \leq i \leq m$. Our goal is to find a mapping $f : X^* \rightarrow Y^*$ that can convert a source sentence to a target sentence sharing the same meaning in the target language (Figure 1). 

\vspace*{-0.5cm}
\ifx\JPicScale\undefined\def\JPicScale{0.5}\fi
\unitlength \JPicScale mm

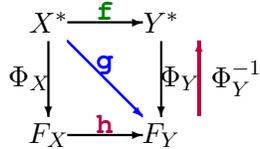
\begin{figure}[ht]
\centering
\begin{picture}(30,40)(27,50)
\put(25,80){\makebox(0,0)[cc]{$X^*$}}

\put(55,80){\makebox(0,0)[cc]{$Y^*$}}

\linethickness{0.3mm}
\put(30,80){\line(1,0){20}}
\put(50,80){\vector(1,0){0.12}}
\linethickness{0.3mm}
\put(25,55){\line(0,1){20}}
\put(25,55){\vector(0,-1){0.12}}
\linethickness{0.4mm}
\multiput(30,75)(0.12,-0.12){167}{\textcolor{blue}{\line(1,0){0.12}}}
\put(50,55){\textcolor{blue}{\vector(1,-1){0.12}}}
\linethickness{0.3mm}
\put(55,55){\line(0,1){20}}
\put(55,55){\vector(0,-1){0.12}}
\linethickness{0.3mm}
\put(30,50){\line(1,0){20}}
\put(50,50){\vector(1,0){0.12}}
\put(25,50){\makebox(0,0)[cc]{$F_X$}}

\put(55,50){\makebox(0,0)[cc]{$F_Y$}}

\put(40,69){\textcolor{blue}{\makebox(0,0)[cc]{\texttt{\textbf{g}}}}}

\put(20,65){\makebox(0,0)[cc]{$\Phi_X$}}

\put(60,65){\makebox(0,0)[cc]{$\Phi_Y$}}

\linethickness{0.4mm}
\put(65,55){\textcolor{purple}{\line(0,1){20}}}
\put(65,75){\textcolor{purple}{\vector(0,1){0.12}}}
\linethickness{0.3mm}
\put(75,65){\makebox(0,0)[cc]{$\Phi_Y^{-1}$}}

\put(40,83){\textcolor{newgreen}{\makebox(0,0)[cc]{\texttt{\textbf{f}}}}}
\put(40,53){\textcolor{purple}{\makebox(0,0)[cc]{\texttt{\textbf{h}}}}}

\end{picture}
\label{s2smap2}
\caption{String-to-string mapping.}
\end{figure}

We define feature mappers $\Phi_X: X^* \rightarrow F_X = \mathbb{R}^{N_X}$ and $\Phi_Y: Y^* \rightarrow F_Y = \mathbb{R}^{N_Y}$ that map each string sequence to a point in high dimensional real number space. 
Let $\textbf{M}_{X} \in \mathbb{R}^{N_X \times m}$ and $\textbf{M}_{Y} \in \mathbb{R}^{N_Y \times m}$ such that $\textbf{M}_{X} = [\Phi_X(\textbf{x}_1), \ldots, \Phi_X(\textbf{x}_m)]$ and $\textbf{M}_{Y} = [\Phi_Y(\textbf{y}_1), \ldots, \Phi_Y(\textbf{y}_m)]$.
The ridge regression solution using $L_2$ regularization is found by minimizing the following cost: 

\vspace*{-0.2cm}
{\small
\begin{equation}
\label{CostRegMT}
\textbf{W}_{L_2} \!= \!\!\!\argmin_{\textbf{W} \in \mathbb{R}^{N_Y \times N_X}} \!\!\!\!\parallel \!\textbf{M}_{Y} - \textbf{W} \textbf{M}_{X} \!\parallel^2_F + \lambda \!\parallel \!\textbf{W}\!\parallel^2_F. \!\!\!
\end{equation}
}


Two main challenges of the regression based machine translation (RegMT) approach are learning the regression function, $h : F_X \rightarrow F_Y$, and solving the \textit{pre-image problem}, which, given the features of the estimated target string sequence, $h(\Phi_X(\textbf{x})) = \Phi_Y(\hat{\textbf{y}})$, attempts to find $\textbf{y} \in Y^*$:
$\textbf{y} = \argmin_{\textbf{y} \in Y^*} || h(\Phi_X(\textbf{x})) - \Phi_Y(\textbf{y})||^2$.

\subsection{$L_1$ Regularized Regression}

String kernels lead to sparse feature representations and $L_1$ regularized regression is effective to find the mappings between sparsely observed features.
We would like to observe only a few nonzero target coefficients corresponding to a source feature in the coefficient matrix. 
$L_1$ regularization helps us achieve solutions close to permutation matrices by increasing sparsity~\cite{PRML}. 


$\textbf{W}_{L_2}$ is not a sparse solution and most of the coefficients remain non-zero. 
We are interested in penalizing the coefficients better; zeroing the irrelevant ones leading to sparsification to obtain a solution that is closer to a permutation matrix. $L_1$ norm behaves both as a feature selection technique and a method for reducing coefficient values.

\vspace*{-0.4cm}
{\small
\begin{eqnarray}
\!\!\!\!\!\!\textbf{W}_{L_1} & \!\!\!\!= & \!\!\!\!\!\!\!\argmin_{\textbf{W} \in \mathbb{R}^{N_Y \times N_X}} \!\!\!\parallel \!\textbf{M}_{Y} - \textbf{W} \textbf{M}_{X} \!\parallel^2_F + \lambda \!\parallel \!\textbf{W}\!\parallel_1.
\label{Lasso}
\end{eqnarray}
}Equation~\ref{Lasso} presents the \textit{lasso}
~\cite{Tibshirani1996Regression} solution where the regularization term is now the $L_1$ matrix norm defined as $\parallel \!\!\! \textbf{W} \!\!\! \parallel_1 = \sum_{i,j} |W_{i,j}|$. $\textbf{W}_{L_2}$ can be found by taking the derivative 
but since $L_1$ regularization cost is not differentiable, $\textbf{W}_{L_1}$ is found by optimization or approximation techniques. We use forward stagewise regression (FSR)~\cite{Hastie06forwardstagewise}, which approximates \textit{lasso} for $L_1$ regularized regression. 

\subsection{Instance Selection}

Proper selection of training instances plays an important role for accurately learning feature mappings with limited computational resources. 
Coverage of the features is important since if we do not have the correct features in the training matrices, we will not be able to translate them. Coverage is measured by the percentage of target features of the test set found in the training set. 
For each test sentence, we pick a limited number of training instances designed to improve the coverage of correct features to build a regression model.
We use a technique that we call \textit{dice}, which optimizes source language bigram coverage such that the difficulty of aligning source and target features is minimized. 
We define Dice's coefficient score as:

\vspace*{-0.3cm}
{\small
\begin{equation}
dice(x,y) = \frac{2 C(x, y)}{C(x) C(y)},
\end{equation}
}where $C(x,y)$ is the number of times $x$ and $y$ co-occurr and $C(x)$ is the count of observing $x$ in the selected training set. Given a test source sentence, $S$, 
we can estimate the goodness of a target sentence, $T$, by the sum of the alignment scores: 

\vspace*{-0.3cm}
{\small
\begin{equation} 
\phi_{dice}(S, T) = \frac{1}{|T| \log |S|}\sum_{x \in X(S)} \sum_{j=1}^{|T|} \sum_{y \in Y(x)} dice(y,T_j), 
\end{equation}
}where $X(S)$ stores the features of $S$ and $Y(x)$ lists the tokens in feature $x$. We assume that tokens are separated by the space character.
The difficulty of word aligning a pair of training sentences, $(S,T)$, can be approximated by $|S|^{|T|}$ as for each source token we have $|T|$ alternatives to map to. We use a normalization factor proportional to $|T| \log |S|$, which corresponds to the $\log$ of this value. 


\section{Regression Experiments}



We experiment with different regression techniques and compare their performance to $L_2$ and \textit{lasso}. We show that \textit{lasso} is able to identify features better than other techniques we compared while making less error. 

We use the German-English (\textit{de-en}) parallel corpus of size about $1.6$ million sentences from the WMT'10~\cite{WMT:2010} to select training instances. For development and test sentences, we select randomly among the sentences whose length is in the range $[10, 20]$ and target language bigram coverage in the range $[0.6, 1]$. We select $20$ random instances from each target coverage decimal fraction (i.e. 20 from $[0.6, 0.7]$, 20 from $[0.7, 0.8]$, etc.) to obtain sets of $100$ sentences. We create in-domain \textit{dev}, \textit{dev2}, and \textit{test} sets following this procedure and making sure that test set source sentences do not have exact matches in the training set. 
We use $n$-spectrum weighted word kernel~\cite{KernelMethods2004} as feature mappers which considers all word sequences up to order $n$:

\vspace*{-0.4cm}
{\small
\begin{equation}
k(\textbf{x}, \textbf{x}^\prime) \!= \!\!\!\sum_{p=1}^n \!\!\sum_{i=1}^{|x|-p+1} \!\sum_{j=1}^{|x^\prime|-p+1} \!\!\!\!\!\!p \; I(\textbf{x}[i\!:\!i+p-1] \!=\! \textbf{x}^\prime[j\!:\!j+p-1])
\label{SpectrumKernel}
\end{equation}
}where $\textbf{x}[i\!\!:\!\!j]$ denotes a substring of $\textbf{x}$ with the words in the range $[i,j]$, $I(.)$ is the indicator function, and $p$ is the number of words in the feature. 
Features used are $1$-grams, $2$-grams, or $1$\&$2$-grams.

%


\subsection{Tuning for Target $F_1$}

We perform parameter optimization for the machine learning models we use to estimate the target vector. 
The model parameters such as the regularization $\lambda$ and iteration number for FSR are optimized on \textit{dev} using the $F_1$ measure over the 0/1-class predictions obtained after thresholding $\Phi_Y(\hat{\textbf{y}})$. Let TP be the true positive, TN the true negative, FP the false positive, and FN the false negative rates, the measures are defined as:

\vspace*{-0.4cm}
{\small
\begin{eqnarray}
\mbox{prec} = \frac{\mbox{TP}}{\mbox{TP} + \mbox{FP}}, & \mbox{BER} = (\frac{\mbox{FP}}{\mbox{TN}+\mbox{FP}} + \frac{\mbox{FN}}{\mbox{TP}+\mbox{FN}}) / 2 \\
\mbox{rec} = \frac{\mbox{TP}}{\mbox{TP} + \mbox{FN}}, & F_1 = \frac{2 \times \mbox{prec} \times \mbox{rec}}{\mbox{prec} + \mbox{rec}}
\end{eqnarray}
}where BER is the balanced error rate, prec is precision, and rec is recall. The evaluation techniques measure the effectiveness of the learning models 
in identifying the features of the target sentence making minimal error to increase the performance of the decoder and its translation quality.

The thresholds are used to map real feature values to $0$/$1$-class predictions and they are also optimized using the $F_1$ measure on \textit{dev}. We compare the performance of $L_2$ regularized ridge regression with $L_1$ regularized \textit{lasso} and $L_1$ regression ($L_1$-reg), where there is no regularization term involved. Then we compare the results we obtain with support vector regression (SVR) using \textit{rbf} kernel and iterative thresholding ($iter$-$\varepsilon$) for $L_1$ minimization~\cite{SparseApproximationThesis}.

%

\begin{table}[ht]
\begin{center}
{\small
\begin{tabular}{lllll}
\hline
& \textit{BER} & \textit{Prec} & \textit{Rec} & \textit{$F_1$} \\ 
\hline
& \multicolumn{4}{c}{\underline{$1$-grams}}\\
$L_2$ & 0.18 &  0.47 &  0.65 &  0.55 \\ 
\textbf{\textit{lasso}} & \textbf{0.15} &  \textbf{0.60} &  \textbf{0.71} &  \textbf{0.65} \\
$L_1$-reg &  0.28 &  0.62 &  0.45 &  0.52 \\
SVR & 0.20 &  0.54 &  0.61 &  0.57 \\ 
$iter$-$\varepsilon$ &  0.17 &  0.40 &  0.68 &  0.50 \\
\cdashline{1-5}
Moses & 0.23 & 0.68 & 0.54 & 0.60 \\
\hline
& \multicolumn{4}{c}{\underline{$2$-grams}}\\
$L_2$ & 0.38 &  0.44 &  0.25 &  0.32 \\
\textbf{\textit{lasso}} & \textbf{0.32} &  \textbf{0.51} &  \textbf{0.37} &  \textbf{0.43} \\
$L_1$-reg &  0.41 &  0.37 &  0.19 &  0.25 \\
SVR & 0.39 &  0.43 &  0.22 &  0.29 \\
$iter$-$\varepsilon$ & 0.41 &  0.60 &  0.18 &  0.27 \\
\cdashline{1-5}
Moses & 0.38 & 0.37 & 0.24 & 0.28 \\
\hline
& \multicolumn{4}{c}{\underline{$1$\&$2$-grams}}\\
$L_2$ & 0.27 &  0.52 &  0.45 &  0.49 \\
\textbf{\textit{lasso}} & \textbf{0.24} &  \textbf{0.57} &  \textbf{0.53} &  \textbf{0.55} \\
\cdashline{1-5}
Moses & 0.30 &  0.58 &  0.40 &  0.47 \\
\hline
\end{tabular}
}\end{center}
\caption{\small Comparison of techniques on \textit{dev2} optimized with $F_1$ value with $100$ instances used for each test sentence using \textit{dice} selection. Dimensions are $N_X \times N_Y \approx 676.62 \times 698.63$ for $1$-grams, $1678.63 \times 1803.85$ for $2$-grams, and $2582.54 \times 2674.02$ for $1$/$2$-grams. Threshold used for Moses is $0.2666$, which is optimized on \textit{dev} set.
}
\label{dev2Results}
\end{table}


The coverage as measured by the percentage of test bigrams found in the training set is $scov$ and $tcov$ for source and target coverage. We measure $scov$ and $tcov$ in \textit{dev2} when using $100$ training instances per test sentence to be ($1.0, 0.96$), ($0.94, 0.74$), and ($0.97, 0.85$) for $1$-grams, $2$-grams, and $1$\&$2$-grams respectively. Table~\ref{dev2Results} present the results on \textit{dev2}, listing BER, prec, rec, and $F_1$ when using $1$-grams, $2$-grams, or both. 
The reason for lower performance with bigrams is likely to be due to lower counts of observing them in the training set. This causes a bigger problem for $L_1$-reg and SVR. 

We compare the performance with Moses's phrase table obtained using all of the parallel corpus with maximum phrase length set to $2$. We obtain Moses target vectors from the target phrase table entries for each source test sentence feature and optimize the threshold for achieving the maximum $F_1$ value on \textit{dev}. This threshold is found to be $0.2666$.
Moses achieves 
$(1.0, 0.98)$, $(0.90, 0.66)$, and $(0.94, 0.92)$ coverage values for $1$-grams, $2$-grams, and $1$\&$2$-grams. These coverage values are higher for $1$-grams and $1$\&$2$-grams, thus Moses retains only likely entries in the phrase table. The results on \textit{dev2} set are given in Table~\ref{dev2Results} separated with dashed lines.

We observe that \textit{lasso} is able to achieve better performance than other techniques in terms of BER, recall and $F_1$ values. $iter$-$\varepsilon$ achieves higher precision in $2$-grams. \textit{lasso} is also better than Moses results in all measures. 

\subsection{Target Feature Estimation as Classification}

We can also interpret the feature mapping problem as a classification problem and estimate whether a feature exists in the target (class 1) or not (class 0). We use logistic regression (logreg) and support vector classification (SVC) to determine the target feature for each feature of the source test sentence. 
When the number of 1's is much smaller than the number of 0's in the training set as we observe in our sparse learning setting, a classifier may tend to choose class 0 over class 1. In such cases, we need to introduce some bias towards choosing 1's. Thresholding approach is helpful in increasing the recall while maintaining a high precision value to optimize F1 measure.
The results on \textit{dev2} is given in Table~\ref{classificationResults}. Our interpretation of feature mapping as a classification problem does not give better results than regression. 

\begin{table}[ht]
\begin{center}
{\small
\begin{tabular}{lllll}
\hline
$1$\&$2$-grams & \textit{BER} & \textit{Prec} & \textit{Rec} & \textit{$F_1$} \\ 
\hline
logreg & 0.33 & 0.58 & 0.34 & 0.43 \\
SVC & 0.30 & 0.47 & 0.41 & 0.44 \\
\hline
\end{tabular}
}\end{center}
\caption{\small Interpreting feature estimation as classification on \textit{dev2}.}
\label{classificationResults}
\end{table}

\subsection{Regression Results}






The regression results on \textit{test} when using $250$ training instances per test sentence is given in Table~\ref{testResults}. $scov$ and $tcov$ is found to be ($1.0, 0.96$), ($0.94, 0.75$), and ($0.96, 0.86$) for $1$-grams, $2$-grams, and $1$\&$2$-grams respectively. We observe that FSR's $F_1$ gains over $L_2$ increase as we used $250$ training instances per test sentence and BER is lower for both. 

\begin{table}[h!]
\begin{center}
{\small
\begin{tabular}{lllll}
\hline
& \textit{BER} & \textit{Prec} & \textit{Rec} & \textit{$F_1$} \\ 
\hline
& \multicolumn{4}{c}{\underline{$1$-grams}}\\
$L_2$ &  0.18 &  0.45 &  0.65 &  0.53 \\
\textbf{\textit{lasso}} & \textbf{0.14} &  \textbf{0.61} &  \textbf{0.73} &  \textbf{0.66} \\
\hline
& \multicolumn{4}{c}{\underline{$2$-grams}}\\
$L_2$ &  0.39 &  0.40 &  0.23 &  0.29 \\
\textbf{\textit{lasso}} & \textbf{0.32} &  \textbf{0.52} &  \textbf{0.37} &  \textbf{0.43} \\
\hline
& \multicolumn{4}{c}{\underline{$1/2$-grams}}\\
$L_2$ &  0.28 &  0.47 &  0.44 &  0.45 \\
\textbf{\textit{lasso}} &  \textbf{0.23} &  \textbf{0.58} &  \textbf{0.54} &  \textbf{0.56} \\ 
\hline
\end{tabular}
}\end{center}
\caption{\small Results on \textit{test} with $250$ instances used for each test sentence using \textit{dice} selection. Dimensions are $N_X \times N_Y \approx 671.33 \times 691.90$ for $1$-grams, $1712.34 \times 1836.18$ for $2$-grams, and $2657.03 \times 2744.96$ for $1$\&$2$-grams. 
}
\label{testResults}
\end{table}

The results show that \textit{lasso} achieves better performance on all feature sets and on all measures when compared with other regression techniques in estimating the target feature for a given source feature.

\section{Graph Decoding Experiments}


We demonstrate machine translation results using graph decoding on the German-English \textit{test} set as well as in a constrained translation domain from Spanish to English (\textit{es-en}) using the categorized EuTrans corpus~\cite{Serrano2009}. The corpus provides a more restricted translation environment for decoding and contains 9000 training, 1000 development, and 3000 test sentences. 

We perform graph-based decoding by first generating a De Bruijn graph from the estimated $\hat{\textbf{y}}$~\cite{CortesS2S2007} and then finding Eulerian paths with maximum path weight. 
We use four features when scoring paths: (1) estimation weight from regression, (2) language model score, (3) brevity penalty as found by $e^{\alpha (l_R - |s| / |path|)}$ for $l_R$ representing the length ratio from the parallel corpus and $|path|$ representing the length of the current path, (4) future cost as in Moses~\cite{Koehn07:moses} and weights are tuned using MERT~\cite{och03:MERT} on the \textit{de-en} \textit{dev} set. 





\begin{figure}[t]
\centering
\hspace*{-0.3cm}
\includegraphics[width=0.36\paperwidth]{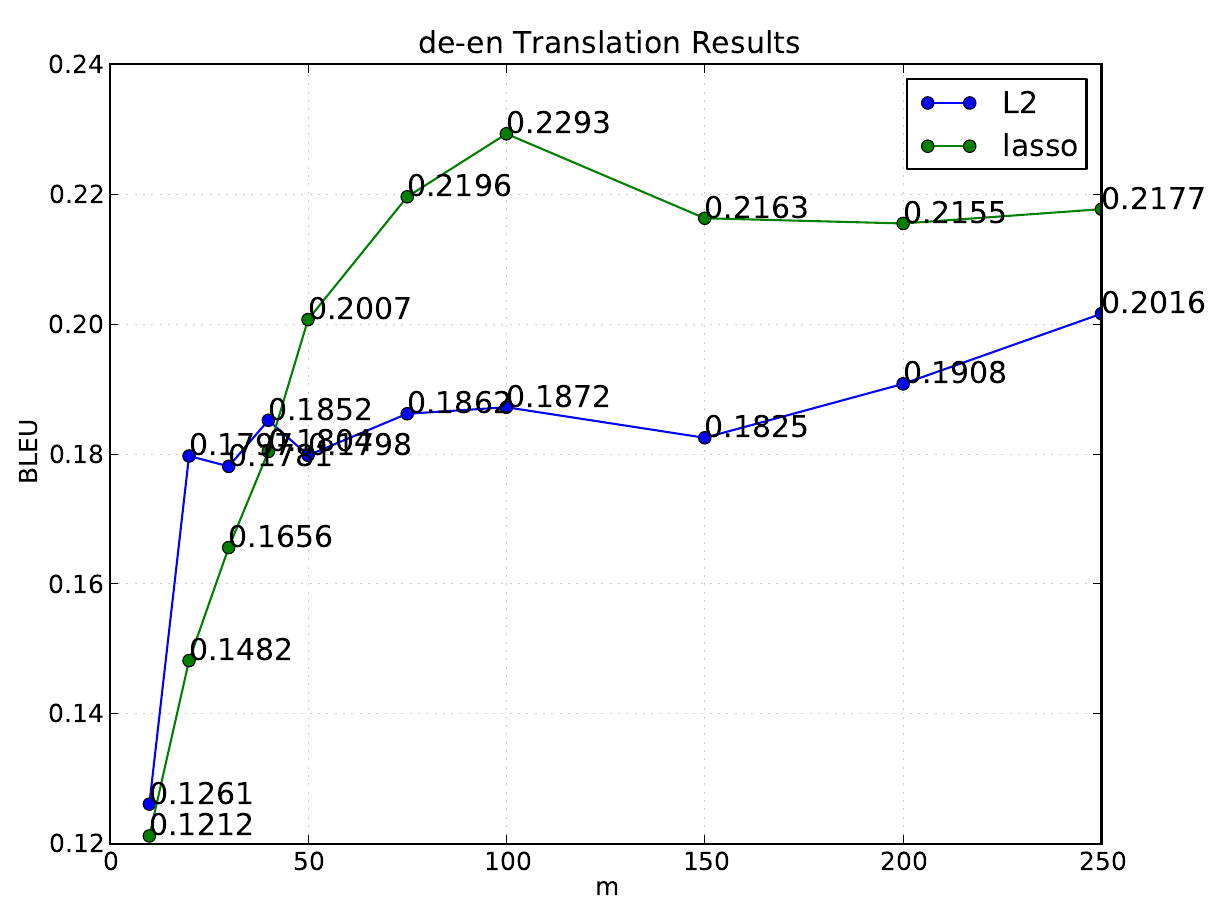}
\caption{\small \textit{de-en} translation results for increasing $m$ using $2$-grams.}
\label{de-enTranslation}
\end{figure}

For \textit{de-en}, we built a Moses model using default settings with maximum sentence length set to $80$ using $5$-gram language model where about $1.6$ million sentences were used for training and $1000$ random development sentences including \textit{dev} used for tuning. We obtained $0.3422$ BLEU score on \textit{test}. Regression results for increasing training data can be seen in Figure~\ref{de-enTranslation} where $2$-grams are used for decoding. We see a large BLEU gain of \textit{lasso} over $L_2$ in our transductive learning setting although the performance is lower than Moses.




%

\begin{figure}[t]
\centering
\hspace*{-0.3cm}
\includegraphics[width=0.36\paperwidth]{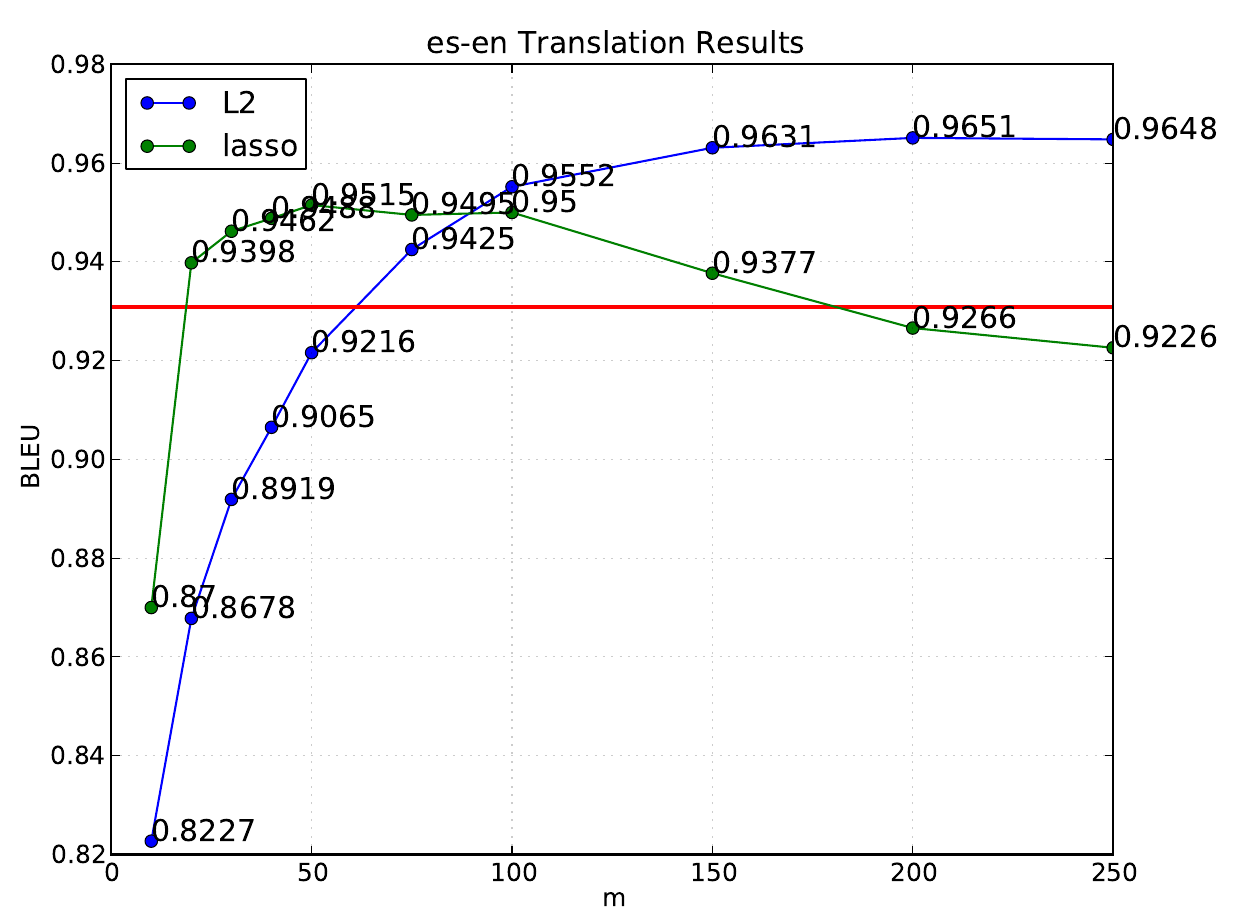}
\caption{\small \textit{es-en} translation results for increasing $m$ using $1$\&$2$-grams.}
\label{es-enTranslationAll}
\end{figure}

For \textit{es-en}, 
Moses achieves $.9340$ BLEU on the full test set.
Regression results for increasing training data can be seen in Figure~\ref{es-enTranslationAll} where $1$\&$2$-grams are used for decoding. We see that \textit{lasso} performs better than $L_2$ in the beginning when we use smaller number of training instances but it performs worse as the training set size increase. The red line corresponds to the Moses baseline. These results are comparable to previous work~\cite{Serrano2009}. 


We demonstrate that sparse $L_1$ regularized regression performs better than $L_2$ regularized regression. Graph based decoding can provide an alternative to state of the art phrase-based decoding system Moses in translation domains with small vocabulary and training set size.

\section{Decoding Experiments with Moses}

Moses uses some extra reordering features, lexical weights, and phrase translation probabilities in both translation directions. 
Therefore we also experiment with decoding using Moses. 

We interpret the coefficient matrix $\textbf{W}$ as the phrase table and perform experiments with Moses using the new phrase table we obtain. These experiments show whether we can effectively use $\textbf{W}$ as the phrase table in a phrase-based decoder. In our transductive learning framework, a $\textbf{W}$ matrix is created for each test sentence and this replaces the phrase table. We modify Moses such that it reloads the phrase table before translating each test sentence. Moses use a default translation table limit (\verb=--ttable-limit=) of $20$ entries per source phrase, which limits the number of translation options considered for each source phrase.

%

We obtain direct and indirect translation probabilities 
for the target phrase entries for each source phrase. We obtain the scores for a given source test sentence S as follows~\footnote{$p(t_f | s_f)$ is the direct phrase translation probability and corresponds to $p(e|f)$ in Moses. Similarly, $p(s_f | t_f)$ is the inverse phrase translation probability and corresponds to $p(f|e)$ in Moses.}:

\vspace*{-0.4cm}
{\small
\begin{eqnarray}
p(t_f | s_f) & = & \frac{\textbf{W}[t_f,s_f]}{\sum_{{t_f}^\prime \in \mathcal{F}(F_Y)} 
 \textbf{W}[{t_f}^\prime,s_f]}, \\
p(s_f | t_f) & = & \frac{\textbf{W}[t_f,s_f]}{\sum_{{s_f}^\prime \in \mathcal{F}(\Phi_X(S))} \textbf{W}[t_f,{s_f}^\prime]}
\end{eqnarray}
}where 
$\mathcal{F}(F_Y)$ is the set of target features found in the training set and $\mathcal{F}(\Phi_X(S))$ return the source features of source test sentence $S$ where $\mathcal{F}(.)$ return the features used in a given domain. 
$s_f$ are chosen from the feature set of the source test sentence and $t_f$ from where $\textbf{W}[t_f,s_f] > 0$. This choice of summing over source features found in the source test sentence helps us discriminate among target feature alternatives better than summing over all source features found in the training set. We also find that by summing over the positive entries for features selected from $\mathcal{F}(F_Y)$ rather than selecting from the top $N$ aligned target features, we increase the precision in the phrase translation scores by increasing the denominator for frequently observed phrases.

We retain the phrase penalty to obtain $3$ scores for each phrase table entry: $p(s_f | t_f), p(t_f | s_f), 2.718$, where the last score is used for the phrase penalty. To compare the performance with Moses, we use the option \verb=-score-options '--NoLex'= during training, which removes the scores coming from lexical weights in phrase table entries, leaving $3$ scores similar to the scores used in the RegMT phrase table. Instead of adding extra lexical weights, we remove it to measure the difference between the phrase tables better.

\subsection{Results}

In our experiments, we measure the effect of phrase length (\verb=--max-phrase-length=), whether to use lexical weights, the number of instances used for training, and the development set selection on the test set performance.
We use \textit{dev} set for tuning the weights, which is constructed similarly to \textit{test} set. We perform individual Moses translation experiment for each test sentence to compare the performance with replacing phrase table with $\textbf{W}$.

For the \textit{de-en} system, we built a Moses model using default settings with maximum sentence length set to $80$ using $5$-gram language model where about $1.6$ million sentences were used for training and $1000$ random development sentences including \textit{dev} used for tuning. We obtained $0.3422$ BLEU score on \textit{test}. 


\textbf{Individual translations:} The training set is composed of only the set of instances selected for each test sentence. Individual translation results obtained with Moses are given in Table~\ref{IndividualMoses}. 
Individual SMT training and translation can be preferable due to smaller computational costs and high parallelizability. 
As we translate a single sentence with each SMT system, tuning weights becomes important and the variance of the weights learned can become high in the individual setting. As we increase the training set size, we observe that the performance gets closer to the Moses system using all of the training corpus.


\begin{table}[ht]
\begin{center}
{\small
\begin{tabular}{llll}
\hline
BLEU & 100 & 250 & 500 \\
\hline
$\leq 2$-grams & 0.3148 & 0.3442 & 0.3709 \\
$\leq 3$-grams & 0.3429 & 0.3534 & 0.4031 \\
\hline
\end{tabular}
}\end{center}
\caption{\small Individual Moses results with training instances selected individually for each source test sentence.
}
\label{IndividualMoses}
\end{table}

\begin{figure}[t]
\centering
\hspace*{-0.3cm}
\includegraphics[width=0.4\paperwidth]{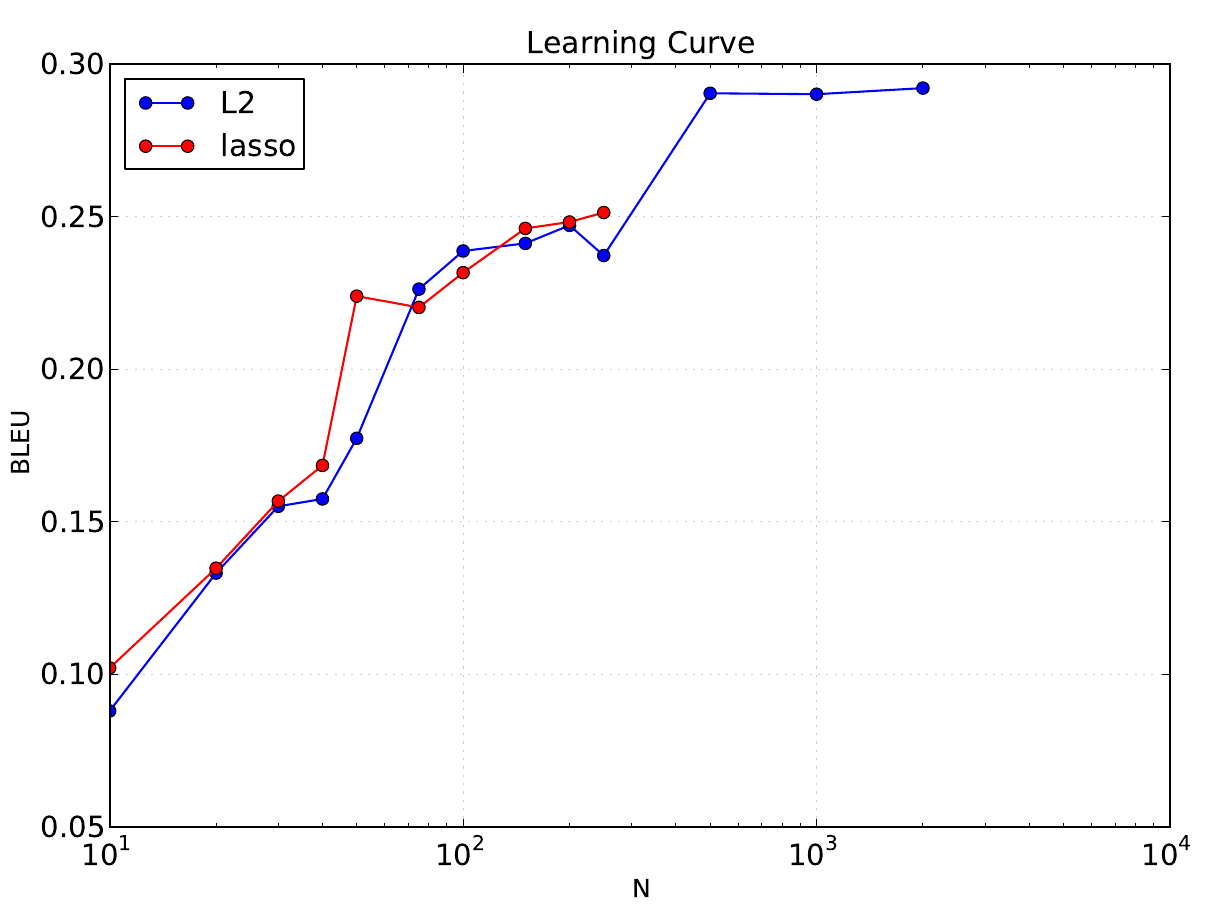}
\caption{\small Moses \textit{de-en} translation results with RegMT \textbf{W} used as the phrase table.}
\label{MosesTranslationAll}
\end{figure}

\textbf{Transductive RegMT \textbf{W} as the phrase table:} The results obtained when the coefficient matrix obtained by RegMT is used as the phrase table for Moses is given in Figure~\ref{MosesTranslationAll}. 
Due to computational limitations, we use the weights obtained for $L_2$ to decode with \textit{lasso} phrase table and skip tuning for \textit{lasso}. 
The learning curve for increasing size of the training set is given in Figure~\ref{MosesTranslationAll}.

We obtain lower performance with $L_2$ when compared with the individual translations obtained. 
\textit{lasso} selects few possible translations for a given source feature. This decrease in vocabulary and testing smaller possibilities may result in lower performance although better estimations are obtained as we observed in the previous section. The increased precision pays when creating the translation from the bigrams found in the estimation. 
We observe similar learning curves both with graph decoding and decoding using Moses. 
RegMT model may need a larger training set size for achieving better performance when the mappings are used as the phrase table. 
RegMT model is good in estimating the target features but has difficulty in correctly finding the target sentence when Moses is used as the decoder.





%

\section{Contributions}

We use transductive regression techniques to learn mappings between source and target features of given parallel corpora and use these mappings to generate machine translation outputs. The results show the effectiveness of using $L_1$ regularization versus $L_2$ used in ridge regression. 
We introduce \textit{dice} instance selection method for proper selection of training instances, which plays an important role to learn correct feature mappings 
We show that $L_1$ regularized regression performs better than $L_2$ regularized regression both in regression measurements and in the translation experiments using graph decoding. We present encouraging results when translating from German to English and Spanish to English.

We also demonstrate results when the phrase table of a phrase-based decoder is replaced with the mappings we find with the regression model. We observe that RegMT model achieves lower performance than Moses system built individually for each test sentence. RegMT model may need a larger training set size for achieving better performance when the mappings are used as the phrase table.

$F_1$ measure is giving us a metric that correlates with BLEU without performing the translation step.

%



\bibliography{machinelearning}
\bibliographystyle{acl}
\end{document}